\documentclass[letterpaper, 10 pt, conference]{ieeeconf}

\IEEEoverridecommandlockouts
\usepackage{graphicx}
\usepackage{xcolor}
\usepackage{amsmath}
\usepackage{amssymb}
\usepackage{booktabs}
\usepackage{multirow}
\usepackage{array}
\usepackage{url}
\usepackage{cite}
\usepackage{flushend}
\usepackage{comment}
\usepackage{kotex}

\makeatletter
\renewcommand{\@IEEEfigurecaptionsepspace}{\vskip 3pt\relax}
\makeatother

\newif\ifshowreviewcomments
\showreviewcommentstrue
\definecolor{reviewBK}{RGB}{165,35,45}
\definecolor{reviewHJ}{RGB}{30,75,165}
\definecolor{reviewYT}{RGB}{20,115,70}

\graphicspath{{fig/}}

\newcommand{\rgbd}{RGB-D}

\title{\LARGE \bf
\textsc{CODA}: Depth-Aligned Scene Completion and Object Decomposition from a Single RGB-D Image}

\usepackage[hidelinks]{hyperref}
\hypersetup{
  pdftitle={CODA: Depth-Aligned Scene Completion and Object Decomposition from a Single RGB-D Image},
  pdfauthor={Dongwon Son, Junhyek Han, Yoontae Cho, Minseok Lee, Hong-seok Choi, Jiwook Choi, Hyungjin Kim, Beomjoon Kim}
}

\author{%
\authorblockN{%
\href{https://dongwon-son.github.io/}{Dongwon Son}$^{1}$\quad
\href{https://junhyekh.github.io/}{Junhyek Han}$^{1}$\quad
Yoontae Cho$^{1}$\quad Minseok Lee$^{1}$\\
Hong-seok Choi$^{2}$\quad Jiwook Choi$^{2}$\quad Hyungjin Kim$^{2}$\quad
\href{https://beomjoonkim.github.io/}{Beomjoon Kim}$^{1}$}
\authorblockA{$^{1}$KAIST\qquad $^{2}$Samsung Heavy Industries Co., Ltd.}}

\begin{document}
\bstctlcite{IEEEBSTcontrol}

\maketitle
\thispagestyle{empty}
\pagestyle{empty}

\begin{abstract}
Robots operating safely in cluttered everyday environments often need to infer scene geometry from partial observations. Methods that detect objects in 2D and reconstruct them independently struggle in such scenes: a missed object is never reconstructed, a merged detection can fuse two objects, and separately reconstructed meshes may overlap or fail to touch their supporting surfaces.
We introduce \textsc{CODA} (Complete Once, Decompose Afterward), a generative model that instead reconstructs the complete scene geometry from a single unsegmented \rgbd{} image, then separates the surface into the surrounding environment and movable objects.
Still, generated scene geometry can drift from the observed partial point cloud.
To reduce this drift, \textsc{CODA} uses two explicit 3D grounding mechanisms to keep reconstructed geometry consistent with observed surfaces while completing unseen regions.
Experiments on HomebrewedDB and our custom cluttered-scene dataset show more accurate reconstructions and a higher fraction of objects remaining in place under simulated gravity than both object-first and scene-first baselines.
\par\noindent Project page: {\urlstyle{same}\url{https://dongwon-son.github.io/coda-project-page/}}
\end{abstract}

\section{Introduction}
\begin{figure}[!t]
\centering
\includegraphics[width=\columnwidth,trim=0 250bp 0 0,clip]{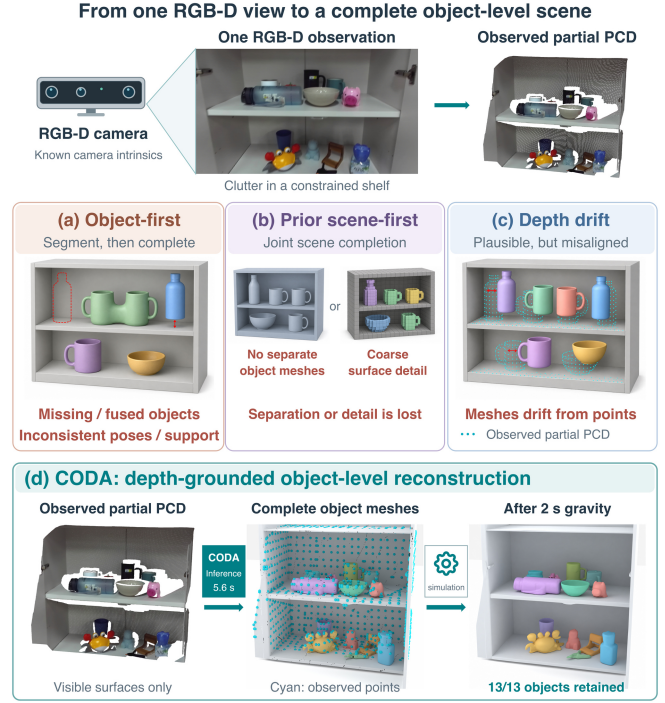}
\caption{\textbf{Reconstruction challenges and \textsc{CODA}.} (a--c) Conceptual illustrations of missing or fused objects, inconsistent poses or support, absent instance separation or fine detail, and drift from observed depth. (d) Real \textsc{CODA} inference converts the observed partial point cloud into complete object meshes. Cyan points show depth grounding. All 13 evaluated objects remain stable over 2\,s in the predicted shelf.}
\label{fig:motivation}
\end{figure}

Practical deployment of the robot in everyday environments requires it to understand cluttered and unstructured environments.
Consider a robot retrieving a target object from a cluttered shelf, such as the one shown in Fig.~\ref{fig:motivation}.
Retrieving the object without collisions requires more than a map of the directly visible surface.
The robot needs complete 3D geometry for the target, neighboring movable objects, and the shelf, all at the correct scale and pose in the observed scene.
A single \rgbd{} observation reveals only a fraction of this geometry, particularly under severe occlusion.
Hidden shape and even object count may be ambiguous, so a useful reconstruction system should remain consistent with measured depth while representing multiple plausible completions.
We study this problem: from one \rgbd{} view with known camera intrinsics (top row of Fig.~\ref{fig:motivation}), reconstruct the complete 3D scene and return separate meshes for the surrounding structure and each movable object.

For this task, there is growing interest in using general-purpose multimodal agents~\cite{anthropic2026fable51,openai2026astra}, which offer a flexible approach: users specify the observations, desired outputs, and task goals without prescribing a reconstruction algorithm.
In our tests, ChatGPT Astra~\cite{openai2026astra} generated scene and object meshes from \rgbd{} observations and camera calibration by writing code to fit visible surfaces to depth measurements and model unseen surfaces.
However, this workflow took approximately 30 minutes per scene in our tests.
Such latency would force long pauses during manipulation when the robot needs an updated reconstruction---for example, after moving an obstructing object to reveal the target before its next grasp.

This latency motivates the use of specialized reconstruction methods, which broadly follow two strategies: \emph{object-first} and \emph{scene-first}.
Object-first approaches first detect and segment visible objects in the input image, then reconstruct each object independently from its masked image and assemble the resulting meshes into a scene.
RecGen~\cite{recgen2026}, ZeroGrasp~\cite{zerograsp2025}, SAM 3D~\cite{sam3d2025}, and SceneComplete~\cite{scenecomplete} follow this pipeline.
These methods therefore depend heavily on the quality of the initial segmentation: missed objects remain absent, while objects merged into one mask may be reconstructed as a single object (Fig.~\ref{fig:motivation}(a)).
Moreover, because each mesh is reconstructed and registered independently, neighboring objects may be misaligned, interpenetrate, or lack valid support (Fig.~\ref{fig:motivation}(a)).

In contrast, scene-first approaches reconstruct the scene jointly rather than reconstructing pre-segmented objects independently.
Joint reconstruction allows these approaches to use neighboring surfaces and supporting structures to infer hidden object shapes and their relative positions.
OctMAE~\cite{iwase2024octmae} and GenRecon~\cite{schmid2026genrecon} reconstruct multiple objects jointly but do not return separate movable-object geometry.
SIMstack~\cite{landgraf2021simstack}, CoReNet~\cite{popov2020corenet}, and PaSCo~\cite{cao2024pasco} provide object- or class-level outputs, but their reconstructions may lack the fine surface detail needed for precise manipulation~\cite{burde2026graspbenchmark,anand2026scout,nolte2025ready} (Fig.~\ref{fig:motivation}(b)).

To reconstruct detailed, separate object meshes without relying on initial 2D segmentation, we propose \textsc{CODA} (\textbf{\underline{C}}omplete \textbf{\underline{O}}nce, \textbf{\underline{D}}ecompose \textbf{\underline{A}}fterward), which completes one detailed scene and separates it into objects only afterward (Fig.~\ref{fig:motivation}(d)).
Reconstructing objects and their supports together helps determine hidden shapes and relative positions, while separating them afterward provides the individual meshes needed for manipulation.
To achieve this, \textsc{CODA} builds on recent advances in image-conditioned 3D generation, which enable detailed geometry and texture generation from a single partial-view image.
Our key idea is to repurpose these pretrained capabilities for scene reconstruction and object separation.
Using TRELLIS.2~\cite{xiang2025trellis2} as our backbone, we fine-tune its geometry flow model~\cite{lipman2023flowmatching} to complete the entire scene and its texture flow model to predict object identities instead of surface appearance.
Because reconstruction and object decomposition both process the whole scene, \textsc{CODA} can complete and separate the scene without a 2D instance mask or known object count.

Completing the scene before separating its objects avoids dependence on initial 2D segmentation, while reusing pretrained 3D generators provides detailed geometry.
However, neither this reconstruction order nor fine-tuning alone guarantees that generated surfaces align with the observed depth, potentially shifting a grasp or invalidating a collision check (Fig.~\ref{fig:motivation}(c)).
In particular, TRELLIS.2's pretrained flow models rely on cross-attention to learn correspondences between RGB features and 3D locations implicitly, without explicitly enforcing consistency with the measured surface.

\begin{figure*}[t]
\centering
\includegraphics[width=\textwidth,trim=0 16bp 0 0,clip]{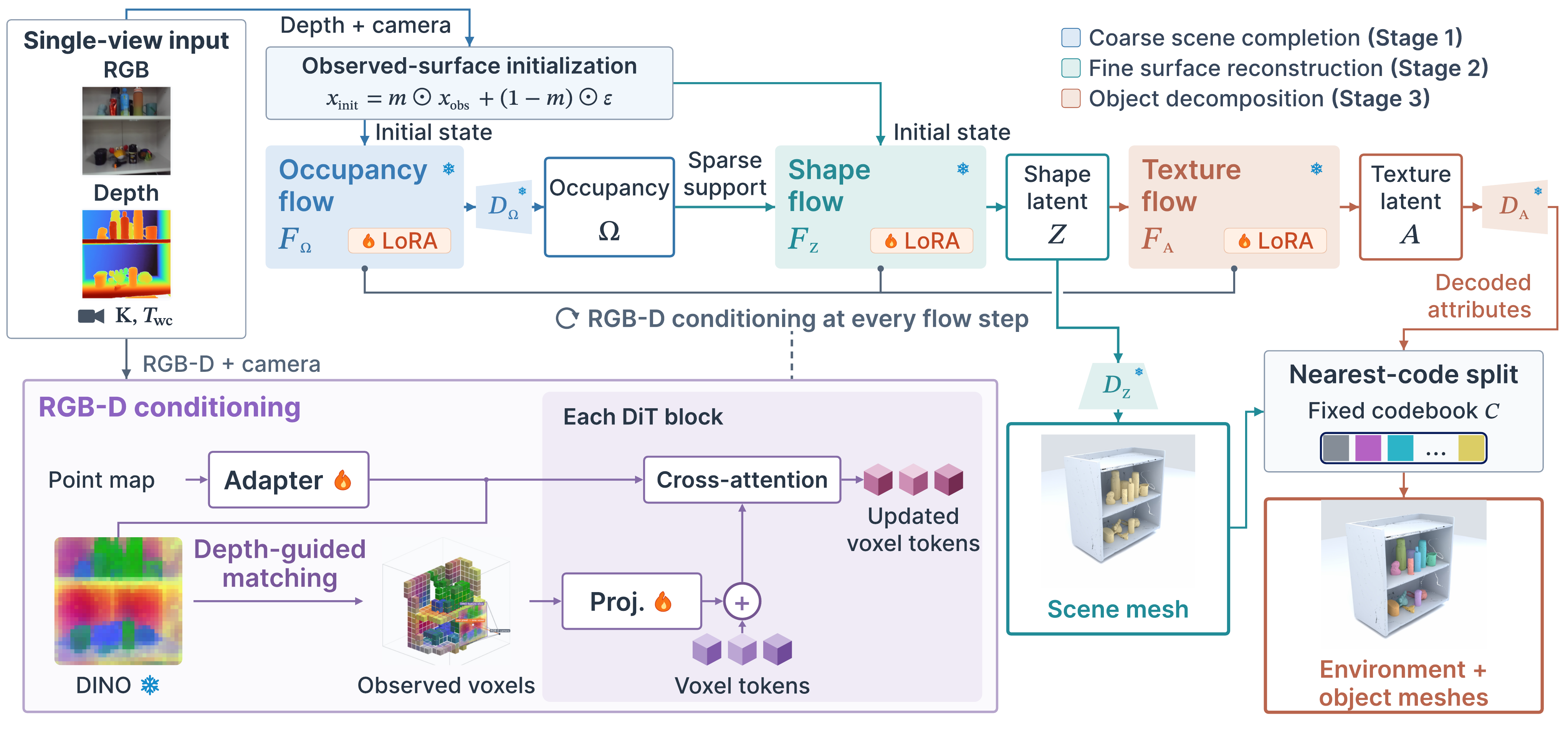}
\caption{\textbf{\textsc{CODA} pipeline.}
Given an unsegmented RGB-D view with camera intrinsics $K$ and
homogeneous camera-to-world transformation $T_{wc}$, Stage~1 generates
coarse occupancy $\Omega$, whose occupied cells determine the
sparse support for Stage~2.
The shape flow generates latent $Z$, which decodes into a scene
mesh.
Stage~3 conditions a pretrained texture flow on $Z$ to generate
latent $A$, repurposing texture attributes as object-identity
codes.
Nearest-code assignment using the fixed codebook $\mathcal{C}$
partitions the scene mesh into environment and object surfaces.
\textbf{Observed-surface initialization (upper left)}
initializes Stages~1 and~2 with encoded partial geometry
$x_{\mathrm{obs}}$ at observed latent sites indicated by $m$,
and Gaussian noise $\epsilon$ elsewhere; Stage~3 starts from
Gaussian noise.
\textbf{RGB-D conditioning (lower left)}
combines DINO image features and point-map features as
cross-attention context.
Depth-guided feature matching additionally back-projects image
features to observed voxels and adds learned projections to
the corresponding voxel tokens before each DiT block in all
three flows.
Each voxel token represents features at one 3D grid location.
$F_{\Omega},F_Z,F_A$ denote the occupancy, shape, and texture flows;
$D_{\Omega},D_Z,D_A$ are their corresponding decoders.
\protect\raisebox{-0.12em}{\protect\includegraphics[height=0.95em]{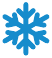}} denotes frozen pretrained weights;
\protect\raisebox{-0.12em}{\protect\includegraphics[height=0.95em]{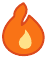}} denotes
trainable adapters and projections.
Output colors indicate object identities.
}
\label{fig:pipeline}
\end{figure*}

To address this misalignment, \textsc{CODA} makes the connection to the RGB-D observation explicit in two complementary ways.
First, it initializes both coarse and fine geometry flow models by combining encoded partial-surface information at observed locations with random noise in unobserved regions, enabling completion much like inpainting in 3D.
Second, it uses camera intrinsics and depth to compute RGB-to-3D correspondences and injects the matched image features alongside the existing cross-attention.
Together, these mechanisms anchor generated surfaces to observed geometry while allowing different completions of hidden regions.
Figure~\ref{fig:pipeline} summarizes the complete flow.

We demonstrate \textsc{CODA} on real-world datasets, including HomebrewedDB~\cite{kaskman2019homebreweddb} and our custom dataset featuring diverse environments and object shapes in cluttered arrangements.
\textsc{CODA} outperforms ChatGPT Astra and the evaluated object-first and scene-first baselines in reconstruction accuracy and pose stability.
Ablation studies further show that our explicit 3D grounding mechanisms improve reconstruction performance.

\section{Related Work}

\subsection{Scene Reconstruction and Object Decomposition}

General-purpose multimodal agents~\cite{anthropic2026fable51,openai2026astra} can choose their reconstruction procedure from the specified observations, camera calibration, and output requirements.
In our tests, ChatGPT Astra used \rgbd{}-guided procedural modeling to construct individual object meshes.

Object-first approaches reconstruct individual objects from object-specific observations or detected object representations~\cite{recgen2026,zerograsp2025,sam3d2025,scenecomplete,kim2025dreamgrasp,wu2026graspfom,yin2026fixer,huang2024zeroshape,centergrasp2024,agnew2021arm,kashyap2025singleview,sucar2020nodeslam,wang2021dspslam,kong2023vmap,chabal2025online}.
Scene-reconstruction pipelines then assemble these object estimates in a common frame.
MIDI~\cite{huang2025midi} and SceneGen~\cite{meng2026scenegen} generate all instances jointly, but they still require a mask for every object.
Task-level benchmarks report frequent collision and stability violations in such reconstructions despite strong vision scores~\cite{burde2026graspbenchmark,anand2026scout,nolte2025ready}.

Scene-first methods instead reconstruct neighboring objects and their support together.
Some methods reconstruct scenes jointly but, unlike \textsc{CODA}, do not produce separate object meshes~\cite{iwase2024octmae,schmid2026genrecon,ren2024scube}.
Others add object separation~\cite{landgraf2021simstack,popov2020corenet,cao2024pasco,dahnert2021panoptic,chu2023buol,cai2021sisnet}.
PaSCo~\cite{cao2024pasco} is particularly relevant to \textsc{CODA}, producing multiple scene-completion hypotheses with instance-level decomposition through an ensemble.
However, its outputs are voxel masks, whereas \textsc{CODA} generates detailed geometry and separates it into object meshes.
SegviGen~\cite{li2026segvigen} repurposes a pretrained texture flow for asset-part segmentation.
\textsc{CODA} adopts this mechanism for scene-level object ownership, assigning the completed surface to the environment and individual objects.
This decomposition is conditioned on both the completed geometry and the original \rgbd{} observation through depth-grounded image-to-3D feature correspondences.

Prior work promotes physical plausibility through stability-aware completion~\cite{agnew2021arm}, equilibrium constraints~\cite{guo2024physcomp,chen2024atlas3d}, or physics-based reconstruction and refinement~\cite{ni2024phyrecon,yao2025cast,ma2026rest3d,xiang2026realtosim}.
\textsc{CODA} jointly reconstructs objects and their surroundings without an explicit physics term, and we evaluate the resulting meshes with a gravity-based settling test~\cite{agnew2021arm,ni2024phyrecon,ma2026rest3d}.
These refinement methods are complementary and could also be applied to \textsc{CODA} outputs.

\subsection{Grounding Generation in Measured Depth}

A single-view reconstruction should stay on the measured surface while generating what the sensor did not see (Fig.~\ref{fig:motivation}(c)).
Two choices govern this: where geometry generation starts, and how image features reach 3D locations.

\textbf{Geometry initialization.}
Mapping systems start from the measurement itself.
Depth fusion~\cite{newcombe2011kinectfusion} and \rgbd{} SLAM with neural fields~\cite{zhu2022niceslam} or 3D Gaussians~\cite{keetha2024splatam} primarily integrate observed surfaces across frames.
Although these methods align reconstructed surfaces with measured depth, they do not target single-view completion of entire objects and scenes, including hidden surfaces.
In contrast, image-conditioned 3D generators such as TRELLIS can produce complete object surfaces, but their geometry may not align with measurements.
To improve alignment, Points-to-3D~\cite{xia2026points3d} initializes the coarse latent from a partial point cloud, while the fine latent still starts from noise.
\textsc{CODA} instead initializes both geometry stages from the observed surface, grounding both the coarse scene structure and fine surface details important for distinguishing nearby objects and preserving support contacts.

\textbf{Image-to-3D correspondence.}
Beyond initialization, reconstruction requires associating image features with 3D locations.
Cross-attention learns these correspondences implicitly.
RecGen~\cite{recgen2026} combines image and depth-derived point-map features through cross-attention, generating an object-centric mesh together with the rotation, translation, and scale needed to place it in the scene.
GenRecon~\cite{schmid2026genrecon} adds explicit spatial correspondence: it projects each voxel into multiple camera views, gathers image features at the projected pixels, and combines them into a feature added directly to that voxel's representation.
\textsc{CODA} adapts this voxel-aligned conditioning to a single \rgbd{} view: measured depth locates image features on the observed 3D surface, and these features supplement cross-attention in all three stages.

\section{Method}

\subsection{Overview}

We consider 3D scene completion and object decomposition from a single unsegmented RGB-D observation.
The input $\mathcal{O}=(I,D,K,T_{wc},\mathcal{B})$ comprises an RGB image $I$, depth map $D$, camera intrinsics $K$, camera-to-world transformation $T_{wc}$, and reconstruction region $\mathcal{B}$.
The world frame is gravity-aligned, with $+z$ opposing gravity.
When $T_{wc}$ is unavailable, we assume an approximate camera-to-world pose can be constructed by placing the world origin at the centroid of the observed point cloud and aligning $+z$ with the upward normal of a fitted support plane assumed to be horizontal.

Figure~\ref{fig:pipeline} summarizes \textsc{CODA}'s three-stage pipeline, which completes a scene from a single \rgbd{} observation before separating it into environment and movable-object meshes.
The Occupancy flow, Shape flow, and Texture flow blocks in its upper row perform coarse scene completion (Stage~1), fine surface reconstruction (Stage~2), and object decomposition (Stage~3), respectively.
Given observation $\mathcal{O}$, \textsc{CODA} samples coarse occupancy $\Omega$, shape latent $Z$, and material latent $A$ according to
\begin{equation}
p(\Omega,Z,A\!\mid\!\mathcal{O})
=p_{\theta}(\Omega\!\mid\!\mathcal{O})\,p_{\phi}(Z\!\mid\!\Omega,\mathcal{O})\,p_{\psi}(A\!\mid\!Z,\mathcal{O}),
\label{eq:factorization}
\end{equation}
where $\theta,\phi,\psi$ parameterize the occupancy, shape, and shape-conditioned texture flows, respectively.
The coarse occupancy $\Omega\in\{0,1\}^{64\times64\times64}$ selects the regions for fine surface reconstruction.
The shape latent $Z$ and material latent $A$ use TRELLIS.2's structured latent (SLat) representation~\cite{xiang2025trellis2}, comprising learned feature vectors at sparse 3D voxel coordinates.
Their feature arrays $Z,A\in\mathbb{R}^{N\times32}$ share $N$ active sites on a grid derived from $\Omega$, with coordinates fixed during sampling.

Conditioned on $\mathcal{O}$, each flow iteratively transforms an initial noisy latent, denoted $x_{\mathrm{init}}^{\Omega}$, $x_{\mathrm{init}}^{Z}$, or $x_{\mathrm{init}}^{A}$ for Stages~1--3, respectively.
Stage~1 generates an occupancy latent that decodes to $\Omega$.
Stage~2 generates $Z$ at the fixed coordinates selected by $\Omega$; the shape decoder then reconstructs a scene mesh at an effective $512^3$ resolution by decoding $Z$.
Stage~3 generates $A$ with additional conditioning on $Z$; the material decoder decodes $A$ into six channels at corresponding surface locations---base-color RGB, metallic, roughness, and opacity---which encode identities used to partition the mesh.
We train only the flow LoRA~\cite{hu2022lora} and additional adapters; all pretrained weights in the flow models, image encoders, and geometry and material encoders and decoders remain frozen.

\subsection{Coarse-to-Fine Scene Reconstruction}
\label{sec:coarse_to_fine}

Both geometry stages use conditional flows~\cite{lipman2023flowmatching} with diffusion transformer (DiT) backbones~\cite{peebles2023dit}, conditioned on RGB and point-map features through cross-attention~\cite{recgen2026}.
We add two mechanisms to ground generation in the observed scene.

\noindent\textbf{Observed-surface initialization.} We extend coarse-only initialization~\cite{xia2026points3d} to both geometry flows.
The point map $P$ uses depth and camera calibration to locate each valid pixel $u$ in 3D world coordinates.
After normalization to $\mathcal{B}$, we encode the observed geometry into voxel-aligned features $x_{\mathrm{obs}}$ for each stage.
Each flow starts from $x_{\mathrm{init}}=m_{\mathrm{obs}}\odot x_{\mathrm{obs}}+(1-m_{\mathrm{obs}})\odot\epsilon$, where the binary mask $m_{\mathrm{obs}}$ marks observed locations and $\epsilon$ is independent standard-normal noise.
Here, $\odot$ denotes element-wise multiplication, with the mask broadcast across feature channels.

\noindent\textbf{Direct image-to-3D feature matching.}
In all three stages, a learned projection reduces the frozen DINO features~\cite{oquab2024dinov2,simeoni2025dinov3}.
We assign each pixel's projected feature to the voxel containing $P(u)$, then average features within each voxel.
Points outside $\mathcal{B}$ or in voxels absent from the active grid are excluded.
Before every DiT block~\cite{peebles2023dit}, a block-specific projection maps these features to the voxel-token width and adds them to the matched tokens.

\noindent\textbf{Optional instance-mask conditioning.} To support the \textsc{CODA} object-first baseline, Stages~1 and~2 can also reconstruct a single object selected by a binary 2D instance mask.
A learned encoder converts the mask into patch tokens added to the RGB-D image features.
Training mixes whole-scene reconstruction without masks and single-object reconstruction with masks.
Standard scene-first \textsc{CODA} disables this mask signal and reconstructs the entire scene.

\noindent\textbf{Training.}
We train the two geometry flows separately with conditional flow matching.
Frozen encoders convert complete ground-truth occupancy and surfaces into target latents for Stages~1 and~2, respectively.
At sampled flow times, the geometry loss is the mean squared error between predicted and target velocities along the interpolation between $x_{\mathrm{init}}$ and the target latent.
Stage~2 uses ground-truth sparse coordinates during training and the support predicted by Stage~1 at inference.

\subsection{Object Decomposition}

To separate the environment and individual objects in the fine scene mesh generated by Stage~2, we adapt the pretrained shape-conditioned texture flow to predict surface identity in Stage~3 (Texture flow block, upper right of Fig.~\ref{fig:pipeline}).

We represent identities using a fixed codebook $\mathcal{C}=\{c_0,\ldots,c_{Q-1}\}$ of $Q$ well-separated vectors $c_j\in[0,1]^6$, where $j$ indexes identity codes.
The codebook size $Q$ includes one environment code $c_0$ and $Q-1$ available movable-object codes; it is not the predicted object count.
For supervision, we label the finest ground-truth surface cells with their environment or object identity and replace each label with its six-channel code using a fixed object ordering.
The frozen material encoder compresses these coded attributes into the target material SLat $A_{\mathrm{gt}}$, whose sparse coordinates match the ground-truth shape SLat $Z_{\mathrm{gt}}$ from the frozen shape encoder.

We train the flow by conditional flow matching from Gaussian noise to $A_{\mathrm{gt}}$, updating LoRA~\cite{hu2022lora} and conditioning adapters while keeping pretrained weights frozen.
At each denoising step, the noisy material latent is concatenated with $Z_{\mathrm{gt}}$ during training and Stage~2's predicted $Z$ at inference.
At inference, the frozen material decoder predicts identity attributes at mesh vertices.
We average each triangle's vertex predictions and assign it to the nearest code in Euclidean distance (Nearest-code split in Fig.~\ref{fig:pipeline}), yielding environment and object meshes without altering geometry.

\section{Experiments}

We evaluate three hypotheses.
H1: \textsc{CODA} outperforms object-first and scene-first baselines in reconstruction accuracy and stability under gravity.
H2: Grounding generation in measured depth improves reconstruction accuracy and agreement with observed surfaces.
H3: Stochastic generation produces diverse hidden shapes while preserving observed geometry.

\subsection{Experimental Setup}

\begin{figure}[!htbp]
\centering
\begin{tabular}{@{}c@{\hspace{2pt}}c@{\hspace{2pt}}c@{\hspace{2pt}}c@{}}
\includegraphics[width=0.24\columnwidth]{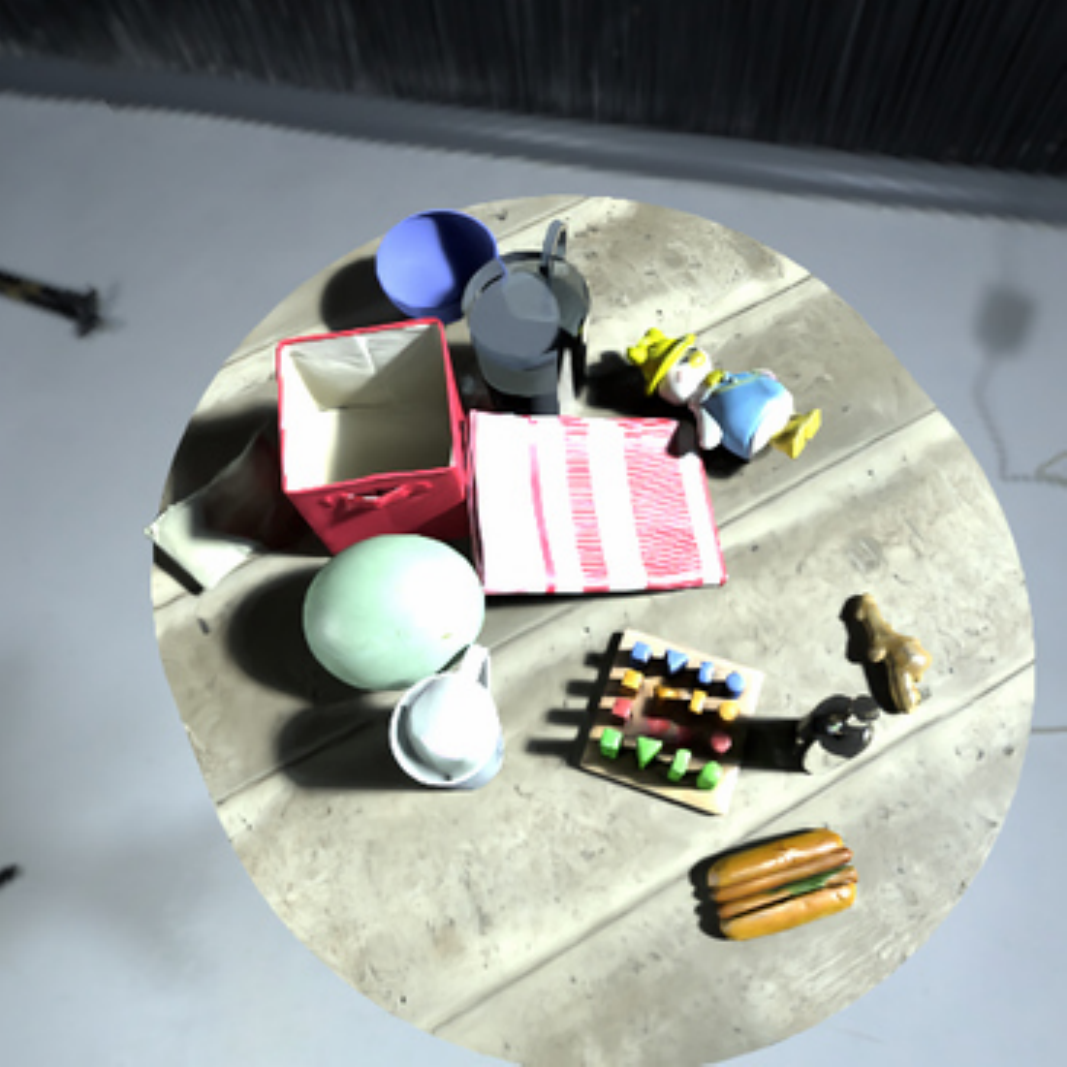} &
\includegraphics[width=0.24\columnwidth]{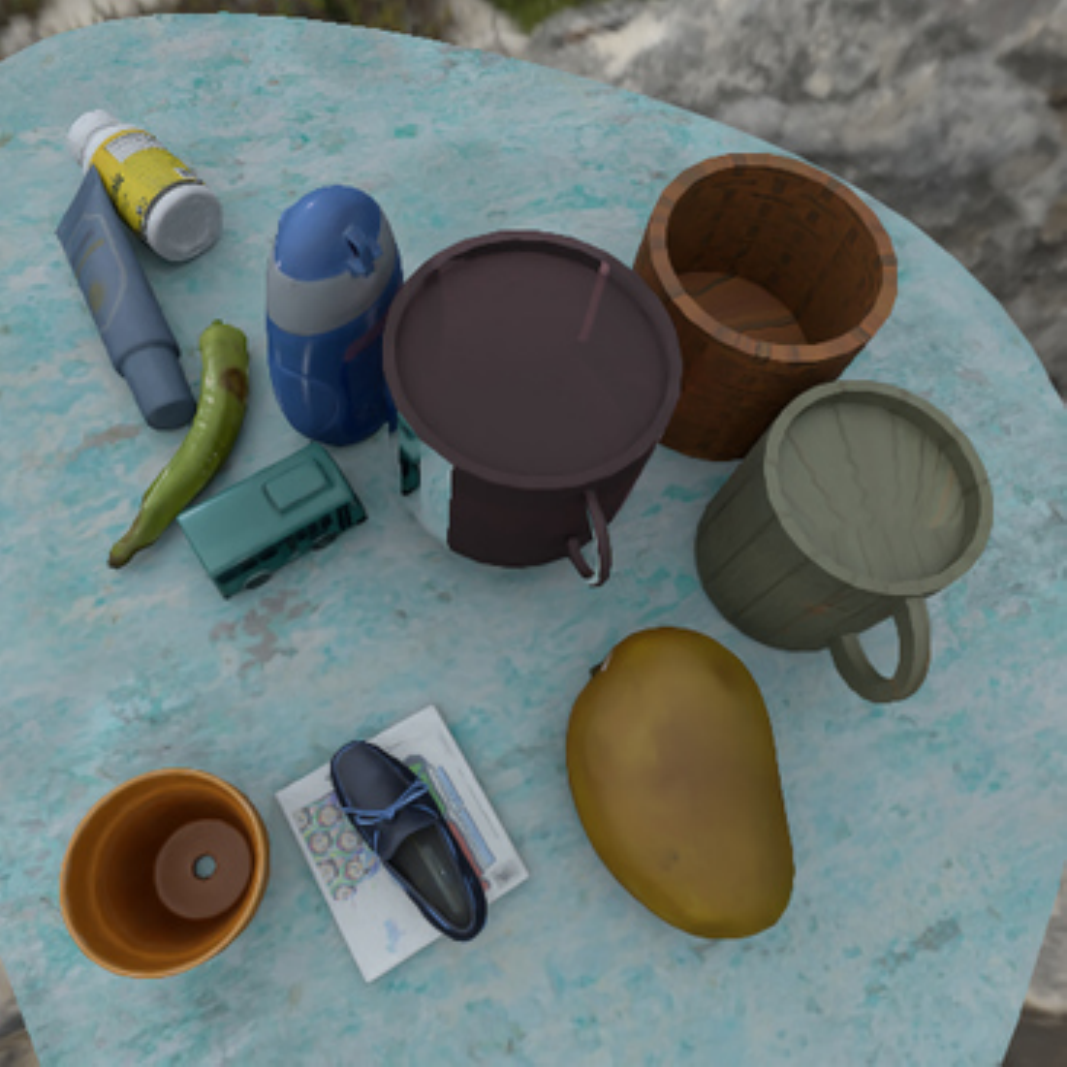} &
\includegraphics[width=0.24\columnwidth]{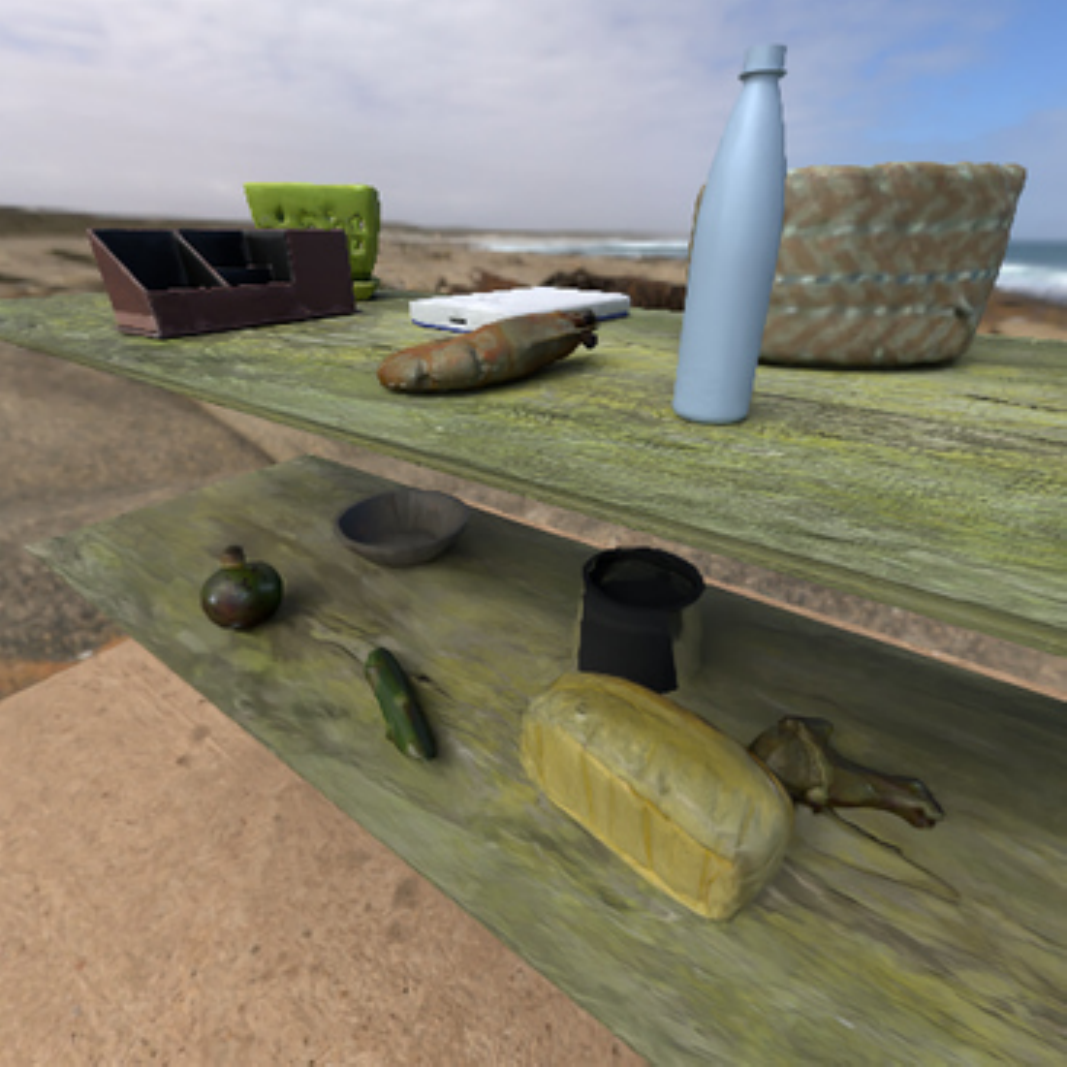} &
\includegraphics[width=0.24\columnwidth]{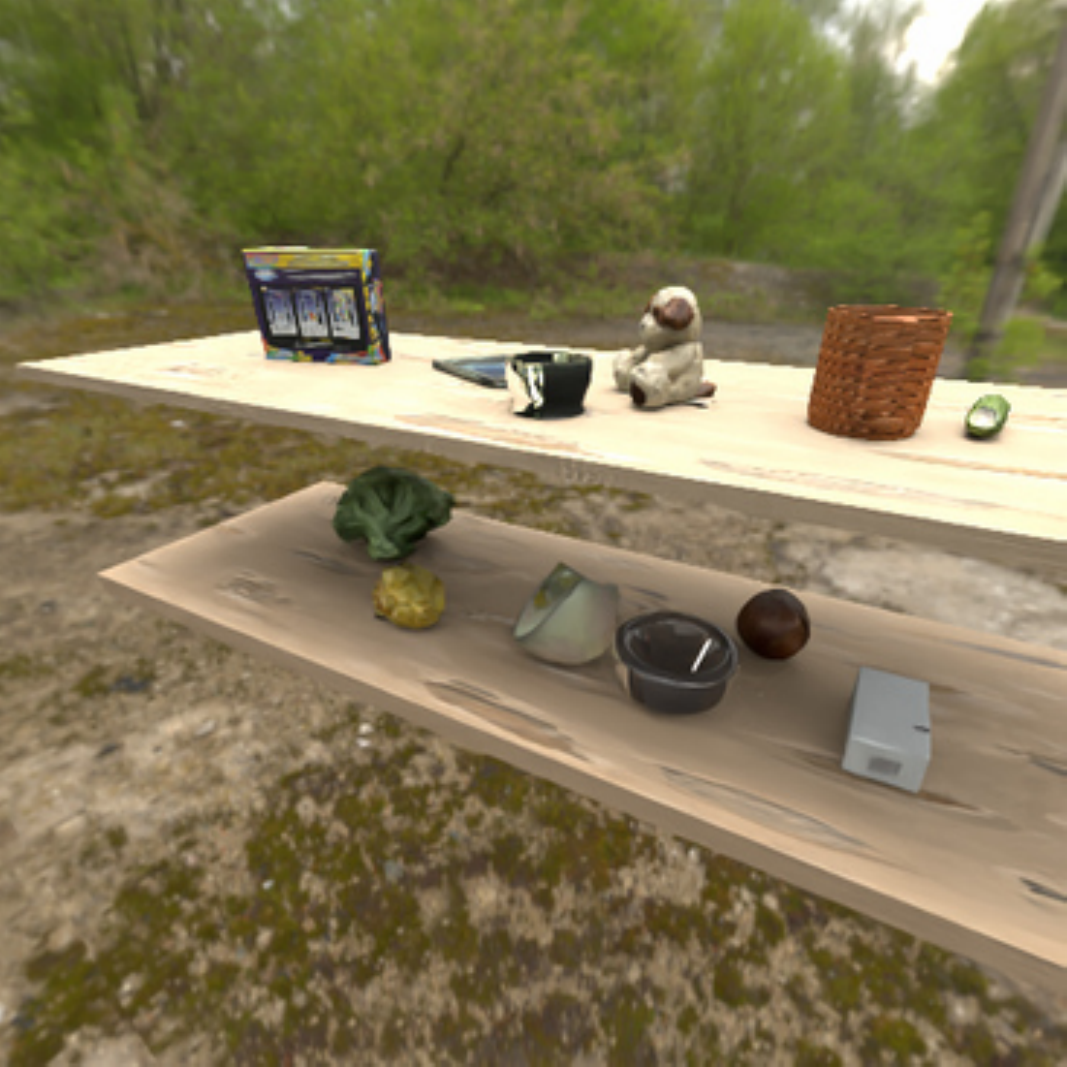}
\end{tabular}
\caption{\textbf{Synthetic dataset examples.} Two table scenes (left) and two shelf scenes (right) from our synthetic dataset.}
\label{fig:dataset_examples}
\end{figure}

Our synthetic dataset is generated entirely in simulation and contains 28K physically settled scenes and 224K \rgbd{} frames, spanning table, shelf, and open-container layouts.
Each scene contains 5--12 objects from Google Scanned Objects~\cite{downs2022gso}, NOCS~\cite{Wang_2019_CVPR_NOCS}, and OmniObject3D~\cite{wu2023omniobject3d}.
MuJoCo~\cite{todorov2012mujoco} supplies settled arrangements and NVISII~\cite{nvisii} renders observations with randomized object pose and scale, camera viewpoint and roll, lighting, and material.
Figure~\ref{fig:dataset_examples} shows cluttered table and shelf examples from our synthetic scene generator.

All methods receive one \rgbd{} view with known camera intrinsics.
For ChatGPT Astra, we specified only the inputs and required output meshes, without prescribing a reconstruction algorithm.
Our \emph{object-first} baselines include RecGen~\cite{recgen2026}, RGB-D-conditioned SAM~3D~\cite{sam3d2025}, ZeroGrasp~\cite{zerograsp2025}, and an object-first \textsc{CODA} variant.
Object-first \textsc{CODA} uses the same Stage~1/2 architecture with optional instance-mask conditioning to reconstruct each object independently, then assembles the meshes into a scene without Stage~3 decomposition.
To examine the effect of input instance segmentation, we evaluate these methods with SAM~3.1~\cite{carion2025sam3} masks and with ground-truth instance masks.

Our \emph{scene-first} baseline is the voxel completion model PaSCo~\cite{cao2024pasco}.
Our PaSCo baseline is based on the original implementation, adapted to our workspace and trained on our synthetic dataset.
\textsc{CODA} and the ChatGPT Astra row in Table~\ref{tab:hb_reconstruction} use raw predicted poses; the other external baselines receive rigid alignment to measured depth using iterative closest point (ICP)~\cite{besl1992registration}.

\begin{figure}[!htbp]
\centering
\includegraphics[width=\columnwidth]{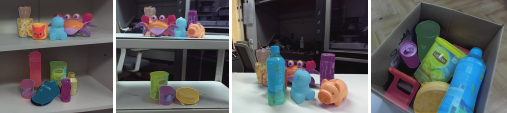}
\caption{\textbf{Custom dataset evaluation examples.} From left: confined shelf, two-tier shelf, table, and container. Semi-transparent instance masks show the projections of registered reference assets onto the evaluated RGB images.}
\label{fig:custom_scene_rgb}
\end{figure}

We use HomebrewedDB~\cite{kaskman2019homebreweddb} for real-world evaluation, but it provides limited coverage of clutter in constrained spaces such as shelves and open containers.
To evaluate these settings, we additionally collect a real-world dataset featuring objects with diverse geometries in cluttered arrangements on tables, on shelves, and inside open containers.
The test set contains 50 RGB-D views across 27 scenes, comprising 449 object-view targets; representative scenes are shown in Fig.~\ref{fig:custom_scene_rgb}.
We will publicly release our collected real-world evaluation dataset after the double-anonymous review process concludes.

\noindent\textbf{Implementation details.}
\textsc{CODA} uses a $1\times1\times1$\,m workspace with $64^3$ coarse occupancy and $512^3$ fine geometry; shape and material SLats have 32 channels.
All three flows use LoRA rank $r=64$ and $\alpha=128$.
We train each stage on one GPU using AdamW with $\beta_1=0.9$, $\beta_2=0.95$, and 1,000 warm-up steps.
Stage~1 uses a base learning rate of $3\times10^{-5}$ with cosine decay to $3\times10^{-6}$ and zero weight decay; Stages~2 and~3 use a constant learning rate of $10^{-4}$ after warm-up and weight decay 0.01.
We evaluate exponential-moving-average weights with decay 0.9999.

\begin{table}[!t]
\centering
\small
\setlength{\tabcolsep}{2pt}
\caption{Reconstruction, pose retention, and runtime. Object-first: SAM~3.1 masks unless marked GT. \textbf{Bold}/\underline{underline}: best/second per group.}
\label{tab:hb_reconstruction}
\begin{tabular*}{\columnwidth}{@{\extracolsep{\fill}}lccc@{}}
\toprule
Method & \shortstack{Pose ret.\\$\uparrow$} & \shortstack{Complete\\F@5 $\uparrow$} & \shortstack{Input view\\F@5 $\uparrow$} \\
\midrule
\multicolumn{4}{l}{\textit{HomebrewedDB}} \\
ChatGPT Astra & \underline{0.7630} & 0.5501 & 0.6877 \\
RecGen & 0.4714 & 0.5929 & 0.8369 \\
SAM3D RGB-D & 0.3571 & 0.5223 & 0.7240 \\
ZeroGrasp & 0.1770 & 0.5674 & \underline{0.9184} \\
PaSCo & 0.1238 & 0.4013 & 0.7906 \\
\textsc{CODA} object-first & 0.6667 & \underline{0.6601} & \underline{0.9184} \\
\textbf{\textsc{CODA} scene} & \textbf{0.7852} & \textbf{0.6789} & \textbf{0.9214} \\
\midrule
\multicolumn{4}{l}{\textit{HomebrewedDB: GT masks}} \\
RecGen & \underline{0.4752} & \underline{0.5997} & 0.8607 \\
SAM3D RGB-D & 0.3465 & 0.5831 & 0.7873 \\
ZeroGrasp & 0.2129 & 0.5882 & \textbf{0.9550} \\
\textsc{CODA} object-first & \textbf{0.6866} & \textbf{0.6719} & \underline{0.9449} \\
\midrule
\multicolumn{4}{l}{\textit{Custom}} \\
RecGen & 0.2584 & 0.6007 & 0.8336 \\
SAM3D RGB-D & 0.2494 & 0.5685 & 0.7899 \\
ZeroGrasp & 0.0713 & 0.5746 & 0.9188 \\
\textsc{CODA} object-first & \underline{0.4922} & \underline{0.6394} & \underline{0.9296} \\
\textbf{\textsc{CODA} scene} & \textbf{0.7661} & \textbf{0.6879} & \textbf{0.9331} \\
\bottomrule
\end{tabular*}
\par\vspace{3pt}
\begin{tabular*}{\columnwidth}{@{\extracolsep{\fill}}cccccc@{}}
\toprule
\multicolumn{6}{c}{\textit{Runtime benchmark: time per scene $\downarrow$}} \\
\midrule
Astra & RecGen & SAM3D & ZeroGrasp & PaSCo & \textbf{\textsc{CODA}} \\
28.1\,min & 100.15\,s & 85.08\,s & 1.54\,s & 2.35\,s & 5.59\,s \\
\bottomrule
\end{tabular*}
\end{table}

\begin{figure}[!t]
\centering
\includegraphics[width=\columnwidth]{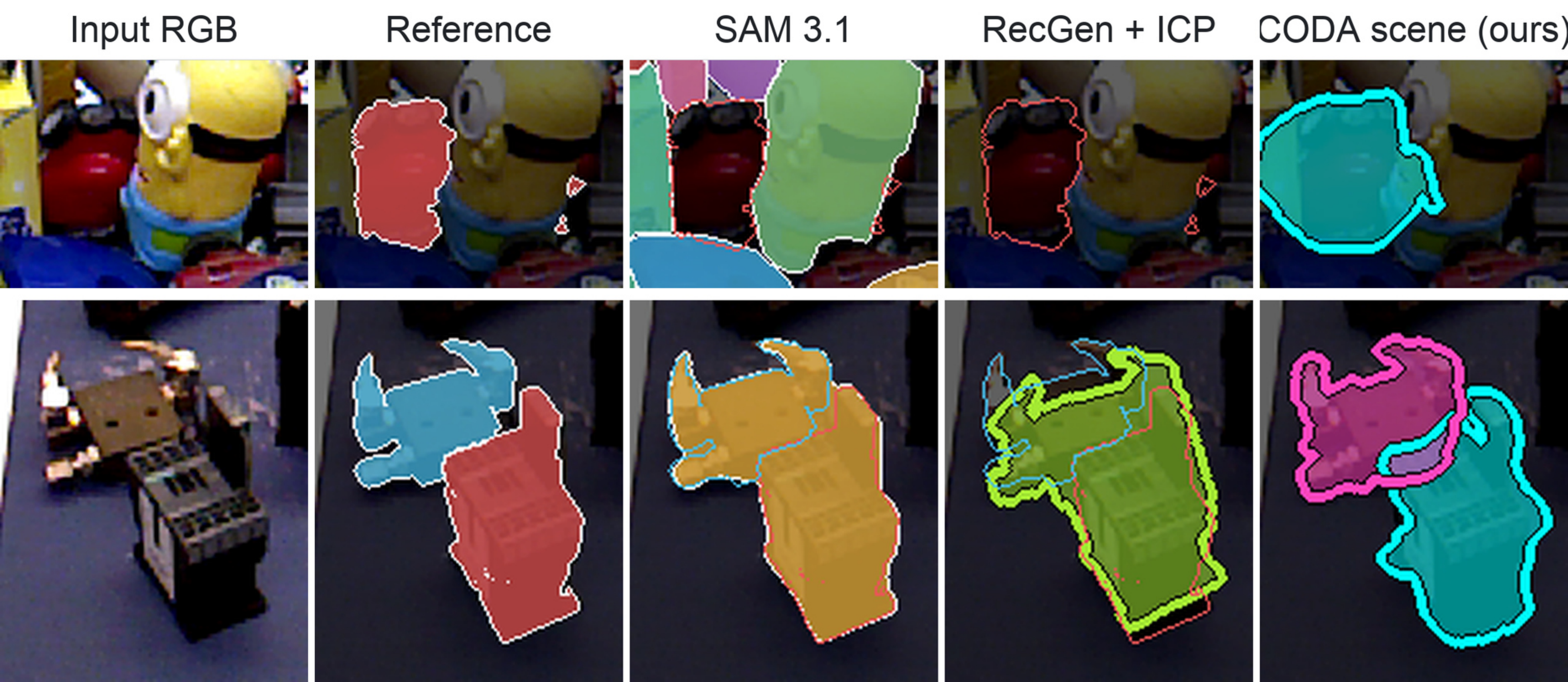}
\caption{\textbf{Representative instance-projection failures with SAM~3.1 proposals on HomebrewedDB.} Top: a severely occluded target has no positive-overlap SAM proposal, so RecGen has no associated reconstruction; \textsc{CODA} scene projects a distinct instance. Bottom: a single SAM proposal covers two annotated objects, and RecGen's projection remains fused, whereas raw-pose \textsc{CODA} scene produces two separate projected instances.}
\label{fig:sam31_coda_instance_failures}
\end{figure}

\begin{figure*}[!t]
\centering
\includegraphics[width=\textwidth]{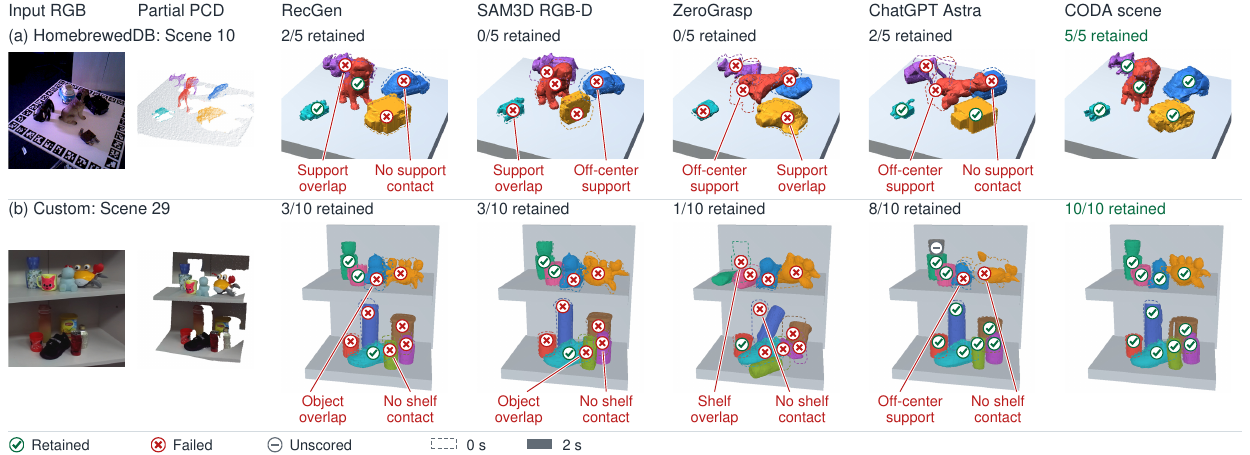}
\caption{\textbf{Object stability under simulated gravity.} Green checks and red crosses mark retained and failed objects at their initial locations. Baselines exhibit initial overlap or inadequate support, whereas \textsc{CODA} retains all evaluated objects in both scenes (5/5 and 10/10) after 2\,s under gravity.}
\label{fig:zed_pose_retention}
\end{figure*}

\subsection{Reconstruction Accuracy, Stability, and Efficiency}
\label{sec:real_scene_reconstruction}

For H1, we report the surface $F_1$-score~\cite{tatarchenko2019what} at a 5\,mm distance threshold (F@5), comparing predictions with reference object geometry (complete surface) or observed surfaces (input view).
HomebrewedDB evaluation covers 39 views and 202 object-view targets.
Hungarian matching associates predictions with reference instances by projected-mask intersection-over-union; zero-overlap pairs are rejected and unmatched targets receive zero F@5.
Complete-surface and input-view F@5 are weighted by target visible-pixel and valid observed-point counts, respectively.
Pose retention is the fraction of evaluated objects satisfying 10\,mm center-of-mass translation and 10-degree rotation limits in a 2\,s gravity simulation.
Higher is better for both metrics.

\noindent\textbf{Results.}
Across both datasets, \textsc{CODA} achieves the highest complete-surface F@5 and pose retention under gravity among the evaluated methods in the main comparison, including both object-first and scene-first baselines.
On HomebrewedDB and our collected dataset, respectively, \textsc{CODA} improves complete-surface F@5 by 14.5\% and 14.5\% relative to RecGen, and input-view F@5 by 0.3\% and 1.6\% relative to ZeroGrasp.
Against ChatGPT Astra on HomebrewedDB, the relative gains are 23.4\% in complete-surface F@5 and 34.0\% in input-view F@5.
\textsc{CODA} achieves retention of 0.785 and 0.766 on HomebrewedDB and our custom dataset, exceeding the strongest external baselines, ChatGPT Astra and RecGen, by 2.2 and 50.8 percentage points, respectively (Table~\ref{tab:hb_reconstruction}).
To assess the effect of reconstruction order, we compare scene-first \textsc{CODA} with its object-first variant on HomebrewedDB.
Scene-first \textsc{CODA} improves complete-surface F@5 by 1.9 percentage points and pose retention by 11.8 percentage points, supporting the effectiveness of reconstructing the scene before separating its objects.
Together, these reconstruction and retention results support H1 across both evaluation datasets.

Qualitatively, we analyze failure cases of object-first methods caused by input instance segmentation: a missed mask leaves an object unreconstructed, while a merged mask can cause two objects to be reconstructed together.
We also tested SAM with object descriptions from a vision-language model, as in SceneComplete~\cite{scenecomplete}, but SAM~3.1 alone with the generic ``individual object'' prompt achieved higher recall in our tests.
Figure~\ref{fig:sam31_coda_instance_failures} illustrates both cases with SAM~3.1: RecGen misses a severely occluded object and merges two objects, whereas scene-first \textsc{CODA} reconstructs distinct instances without input masks.
Furthermore, even when ground-truth masks eliminate input segmentation errors for the object-first baselines, mask-free scene-first \textsc{CODA} achieves higher complete-surface F@5 and pose retention than every GT-mask baseline on HomebrewedDB (Table~\ref{tab:hb_reconstruction}).
Figure~\ref{fig:zed_pose_retention} illustrates gravity retention in two selected real-world scenes, including ChatGPT Astra and the other visualized baselines, with shared depth-derived supports.

\noindent\textbf{Runtime.}\label{sec:runtime}
All runtimes measure steady-state processing from RGB-D input to output meshes, excluding warm-up.
On six custom views evaluated on an RTX~4090, \textsc{CODA} takes 5.59\,s per scene, compared with 100.15\,s for RecGen and 85.08\,s for SAM3D, corresponding to approximately $17.9\times$ and $15.2\times$ speedups.
\textsc{CODA} avoids repeated per-object reconstruction by completing the scene jointly, whereas RecGen and SAM3D reconstruct each object independently.
PaSCo and ZeroGrasp are faster (2.35\,s and 1.54\,s), but achieve lower complete-surface accuracy and pose retention (Table~\ref{tab:hb_reconstruction}).
Astra took 28.1\,min per scene, substantially longer than the reported runtimes of the specialized methods evaluated here.

\subsection{Grounding and Object-Decomposition Ablations}
\label{sec:observed_depth_ablations}
To test H2, Table~\ref{tab:hb_ablations} ablates observed-surface initialization and direct image-to-3D feature matching, including GenRecon-style~\cite{schmid2026genrecon} and Points-to-3D-style~\cite{xia2026points3d} configurations.
Matching ablations apply to all three stages.
The TRELLIS.2 baseline is fine-tuned for RGB-D conditioning.

\begin{table}[!t]
\centering
\small
\setlength{\tabcolsep}{2pt}
\caption{HomebrewedDB ablations. I: initialization; M: matching.}
\label{tab:hb_ablations}
\footnotesize
\setlength{\tabcolsep}{1pt}
\begin{tabular*}{\columnwidth}{@{\extracolsep{\fill}}l*{7}{c}@{}}
\toprule
Variant & \multicolumn{2}{c}{Stage~1} & \multicolumn{2}{c}{Stage~2} & Texture & Complete & Input view \\
\cmidrule(lr){2-3}\cmidrule(lr){4-5}
& I & M & I & M & prior & F@5 $\uparrow$ & F@5 $\uparrow$ \\
\midrule
Full model & $\checkmark$ & $\checkmark$ & $\checkmark$ & $\checkmark$ & $\checkmark$ & \textbf{0.6789} & \textbf{0.9214} \\
Learned-query masks & $\checkmark$ & $\checkmark$ & $\checkmark$ & $\checkmark$ & -- & 0.3748 & 0.6526 \\
No matching & $\checkmark$ & -- & $\checkmark$ & -- & $\checkmark$ & \underline{0.6325} & \underline{0.9085} \\
GenRecon-style & -- & $\checkmark$ & -- & $\checkmark$ & $\checkmark$ & 0.4501 & 0.6489 \\
Points-to-3D-style & $\checkmark$ & -- & -- & -- & $\checkmark$ & 0.3993 & 0.5704 \\
TRELLIS.2 fine-tuned & -- & -- & -- & -- & $\checkmark$ & 0.1396 & 0.1411 \\
\bottomrule
\end{tabular*}
\end{table}

\noindent\textbf{Results.}
The full model achieves complete-surface F@5 of 0.679, compared with 0.633 without matching and 0.450 without initialization.
Coarse-only initialization reaches 0.399, while disabling both mechanisms yields 0.140.
These trends support combining observed-surface initialization with feature matching to improve reconstruction accuracy.
For object decomposition, we replace the pretrained texture flow with a learned-query decoder following PaSCo/Mask2Former~\cite{cao2024pasco,cheng2022mask2former}, without texture-flow pretraining.
The decoder partitions the same saved \textsc{CODA} geometry.
The learned-query variant reaches complete-surface F@5 of 0.3748 and pose retention of 0.2010.
Most failures stem from incorrect separation of objects from the environment.

\begin{table}[!t]
\centering
\small
\setlength{\tabcolsep}{2pt}
\caption{Object-wise YCB reconstruction and diversity (eight seeds).}
\label{tab:occlusion_diversity}
\begin{tabular*}{\columnwidth}{@{\extracolsep{\fill}}llcccc@{}}
\toprule
Group & Method & \shortstack{Best\\F@10} & \shortstack{Visible\\F@10} & $D_{\mathrm{vis}}\downarrow$ & $D_{\mathrm{hid}}\uparrow$ \\
\midrule
\multirow{2}{*}{Easy} & RecGen & 0.8337 & 0.9040 & 0.3564 & 0.5258 \\
& \textbf{\textsc{CODA} scene} & 0.9667 & 0.9769 & 0.0681 & 0.3469 \\
\multirow{2}{*}{Mid} & RecGen & 0.6843 & 0.8140 & 0.4902 & 0.6693 \\
& \textbf{\textsc{CODA} scene} & 0.8899 & 0.9452 & 0.1326 & 0.5073 \\
\multirow{2}{*}{Hard} & RecGen & 0.5172 & 0.7062 & 0.5890 & 0.7461 \\
& \textbf{\textsc{CODA} scene} & 0.7263 & 0.8744 & 0.1724 & 0.6192 \\
\bottomrule
\end{tabular*}
\end{table}

\subsection{Occlusion Accuracy and Object-Wise Diversity}
\label{sec:occlusion_diversity}

To test whether stochastic generation produces diverse hidden shapes while preserving observed geometry (H3), we compare RecGen and \textsc{CODA} by generating eight reconstructions of each target object from the same RGB-D input using fixed random seeds.
We measure reconstruction accuracy and variation in visible and hidden regions as occlusion increases.
We evaluate YCB objects in table and shelf scenes.

We group targets by how much of each object's image silhouette remains visible compared with an unobstructed view from the same camera.
The groups are \textbf{easy} (75--100\% visible), \textbf{mid} (40\% to less than 75\% visible), and \textbf{hard} (10\% to less than 40\% visible).
Table~\ref{tab:occlusion_diversity} reports results for 274 object instances: 91 easy, 90 mid, and 93 hard.

\textbf{Best F@10} ($F^*$) is the highest ground-truth surface F-score among eight samples; \textbf{Visible F@10} ($F_v$) is the mean F-score against measured RGB-D points across samples. Both use a 10\,mm tolerance, with higher values preferred; best-sample selection uses ground truth.
\textbf{Visible and hidden diversity} ($D_{\mathrm{vis}}$, $D_{\mathrm{hid}}$) measure mean pairwise voxel disagreement (one minus intersection-over-union) in observed and unobserved regions, respectively. Lower $D_{\mathrm{vis}}$ indicates observation consistency; higher $D_{\mathrm{hid}}$ indicates varied completions, provided reconstruction quality remains high.
All metrics are averaged equally over objects within each occlusion group.

\begin{figure}[!htbp]
\centering
\includegraphics[width=\columnwidth]{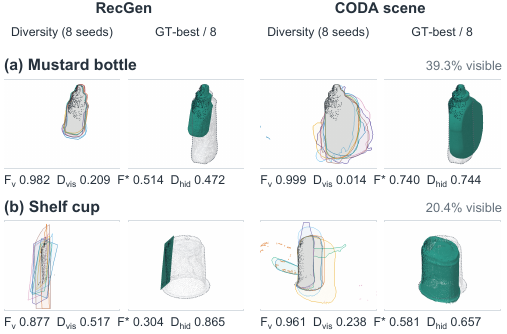}
\caption{Stochastic completion of two hard-occluded YCB objects in simulation. For each method, colored contours show all eight fixed-seed predictions over observed surface points (black), while the paired view shows the oracle GT-best sample (teal) against the ground-truth silhouette (dashed gray).}
\label{fig:object_stochasticity}
\end{figure}

\noindent\textbf{Results.}
Table~\ref{tab:occlusion_diversity} shows higher best-sample and visible F@10 for \textsc{CODA} in every occlusion group.
Relative to RecGen, the best-sample F@10 gain increases from 16.0\% for easy objects to 40.4\% for hard objects.
In the hard group, visible F@10 improves by 23.8\% and $D_{\mathrm{vis}}$ decreases by 70.7\%, indicating more accurate and consistent observed surfaces across samples.
\textsc{CODA}'s hidden-region diversity increases from 0.347 to 0.619 as occlusion increases from easy to hard.
Although RecGen has higher $D_{\mathrm{hid}}$, it also has greater visible-region variation and lower reconstruction accuracy; diversity alone does not imply better completion.
These results support H3: \textsc{CODA} generates varied hidden shapes while maintaining stronger agreement with observed geometry.
Figure~\ref{fig:object_stochasticity} illustrates this behavior on two hard-occluded table and shelf objects.

\section{Conclusion}

We presented \textsc{CODA}, a scene-first approach that reconstructs a complete scene from a single RGB-D observation before decomposing it into environment and object meshes, without input instance masks.
Observed-surface initialization and depth-grounded feature matching connect generative shape priors to measured geometry, while texture-flow-based decomposition recovers individual objects from the completed scene.
Experiments on HomebrewedDB and our collected cluttered scenes show improved reconstruction accuracy and pose stability over the evaluated baselines.
Ablations support the role of explicit depth grounding, and stochastic reconstruction produces varied hidden shapes while maintaining greater consistency in observed regions.
Together, these results support completing the scene before separating its objects as an effective approach to reconstruction under clutter and occlusion.
Future work will apply \textsc{CODA} to object manipulation tasks.

\bibliographystyle{IEEEtran}
\bibliography{ref}

@article{xiang2025trellis2,
  title={Native and Compact Structured Latents for {3D} Generation},
  author={Xiang, Jianfeng and Chen, Xiaoxue and Xu, Sicheng and others},
  journal={Technical report},
  year={2025},
  url={https://github.com/microsoft/TRELLIS.2}
}

@inproceedings{lipman2023flowmatching,
  title={Flow Matching for Generative Modeling},
  author={Lipman, Yaron and Chen, Ricky T. Q. and Ben-Hamu, Heli and Nickel, Maximilian and Le, Matt},
  booktitle={Proc. Int. Conf. Learn. Represent.},
  year={2023},
  url={https://openreview.net/forum?id=PqvMRDCJT9t}
}

@inproceedings{peebles2023dit,
  title={Scalable Diffusion Models with Transformers},
  author={Peebles, William and Xie, Saining},
  booktitle={Proc. IEEE/CVF Int. Conf. Comput. Vis.},
  pages={4195--4205},
  year={2023},
}

@article{schmid2026genrecon,
  title={{GenRecon}: Bridging Generative Priors for Multi-View {3D} Scene Reconstruction},
  author={Schmid, Katharina and von L{\"u}tzow, Nicolas and Hladk{\'y}, Jozef and Dai, Angela and Nie{\ss}ner, Matthias},
  journal={arXiv:2605.23888},
  year={2026},
}

@article{li2026segvigen,
  title={{SegviGen}: Repurposing {3D} Generative Model for Part Segmentation},
  author={Li, Lin and Feng, Haoran and Huang, Zehuan and Chen, Haohua and Nie, Wenbo and Hou, Shaohua and Fan, Keqing and Hu, Pan and Wang, Sheng and Li, Buyu and Sheng, Lu},
  journal={ACM Trans. Graph.},
  volume={45},
  number={4},
  articleno={68},
  pages={1--12},
  year={2026},
  month=jul,
  doi={10.1145/3811399}
}

@inproceedings{xia2026points3d,
  title={{Points-to-3D}: Structure-Aware {3D} Generation with Point Cloud Priors},
  author={Xia, Jiatong and Duan, Zicheng and van den Hengel, Anton and Liu, Lingqiao},
  booktitle={Proc. IEEE/CVF Conf. Comput. Vis. Pattern Recognit.},
  year={2026}
}

@inproceedings{landgraf2021simstack,
  title={{SIMstack}: A Generative Shape and Instance Model for Unordered Object Stacks},
  author={Landgraf, Zoe and Scona, Raluca and Laidlow, Tristan and James, Stephen and Leutenegger, Stefan and Davison, Andrew J.},
  booktitle={Proc. IEEE/CVF Int. Conf. Comput. Vis.},
  pages={13012--13022},
  year={2021}
}

@inproceedings{agnew2021arm,
  title={Amodal {3D} Reconstruction for Robotic Manipulation via Stability and Connectivity},
  author={Agnew, William and Xie, Christopher and Walsman, Aaron and Murad, Octavian and Wang, Caelen and Domingos, Pedro and Srinivasa, Siddhartha},
  booktitle={Proc. 2020 Conf. Robot Learn.},
  series={Proc. Mach. Learn. Res.},
  volume={155},
  pages={1498--1508},
  year={2021},
  publisher={PMLR}
}

@inproceedings{popov2020corenet,
  title={{CoReNet}: Coherent {3D} Scene Reconstruction from a Single {RGB} Image},
  author={Popov, Stefan and Bauszat, Pablo and Ferrari, Vittorio},
  booktitle={Proc. Eur. Conf. Comput. Vis.},
  year={2020}
}

@inproceedings{cai2021sisnet,
  title={Semantic Scene Completion via Integrating Instances and Scene In-the-Loop},
  author={Cai, Yingjie and Chen, Xuesong and Zhang, Chao and Lin, Kwan-Yee and Wang, Xiaogang and Li, Hongsheng},
  booktitle={Proc. IEEE/CVF Conf. Comput. Vis. Pattern Recognit.},
  pages={324--333},
  year={2021}
}

@inproceedings{dahnert2021panoptic,
  title={Panoptic {3D} Scene Reconstruction from a Single {RGB} Image},
  author={Dahnert, Manuel and Hou, Ji and Niessner, Matthias and Dai, Angela},
  booktitle={Adv. Neural Inf. Process. Syst.},
  year={2021}
}

@inproceedings{chu2023buol,
  title={{BUOL}: A Bottom-Up Framework with Occupancy-Aware Lifting for Panoptic {3D} Scene Reconstruction from a Single Image},
  author={Chu, Tao and Zhang, Pan and Liu, Qiong and Wang, Jiaqi},
  booktitle={Proc. IEEE/CVF Conf. Comput. Vis. Pattern Recognit.},
  pages={4937--4946},
  year={2023}
}

@inproceedings{iwase2024octmae,
  title={Zero-Shot Multi-Object Scene Completion},
  author={Iwase, Shun and Liu, Katherine and Guizilini, Vitor and others},
  booktitle={Proc. Eur. Conf. Comput. Vis.},
  year={2024}
}

@inproceedings{ren2024scube,
  title={{SCube}: Instant Large-Scale Scene Reconstruction using {VoxSplats}},
  author={Ren, Xuanchi and Lu, Yifan and Liang, Hanxue and Wu, Zhangjie and Ling, Huan and Chen, Mike and Fidler, Sanja and Williams, Francis and Huang, Jiahui},
  booktitle={Adv. Neural Inf. Process. Syst.},
  year={2024}
}

@inproceedings{cao2024pasco,
  title={{PaSCo}: Urban {3D} Panoptic Scene Completion with Uncertainty Awareness},
  author={Cao, Anh-Quan and Dai, Angela and de Charette, Raoul},
  booktitle={Proc. IEEE/CVF Conf. Comput. Vis. Pattern Recognit.},
  year={2024}
}

@inproceedings{cheng2022mask2former,
  title={Masked-attention Mask Transformer for Universal Image Segmentation},
  author={Cheng, Bowen and Misra, Ishan and Schwing, Alexander G. and Kirillov, Alexander and Girdhar, Rohit},
  booktitle={Proc. IEEE/CVF Conf. Comput. Vis. Pattern Recognit.},
  pages={1290--1299},
  year={2022}
}

@article{kim2025dreamgrasp,
  title={Zero-Shot Scene Geometry Recognition from Sparse Partial-View Images: Leveraging Generative Diffusion Models for Robotic Manipulation Tasks},
  author={Kim, Young Hun and Kim, Seungyeon and Lee, Yonghyeon and Park, Frank Chongwoo},
  journal={IEEE Robot. Autom. Lett.},
  year={2026}
}

@article{wu2026graspfom,
  title={{GraspFoM}: Towards Reconstruction-Driven Robotic Grasping with {3D} Foundation Priors},
  author={Wu, Dongli and Wei, Xiaobao and Wang, Hao and Dong, Qiaochu and Li, Ying and Wuwu, Qingpo and Lu, Ming and Zhao, Wufan},
  journal={arXiv:2606.08440},
  year={2026}
}

@inproceedings{yin2026fixer,
  title={{3D-Fixer}: Coarse-to-Fine In-place Completion for {3D} Scenes from a Single Image},
  author={Yin, Ze-Xin and Liu, Liu and Wang, Xinjie and Sui, Wei and Su, Zhizhong and Yang, Jian and Xie, Jin},
  booktitle={Proc. IEEE/CVF Conf. Comput. Vis. Pattern Recognit.},
  pages={12753--12763},
  year={2026}
}

@inproceedings{ma2026rest3d,
  title={{REST3D}: Reconstructing Physically Stable {3D} Scenes from a Single Image},
  author={Ma, Xiaoxuan and Wang, Jiashun and Ugrinovic, Nicolas and Litman, Yehonathan and Kitani, Kris},
  booktitle={ACM SIGGRAPH Asia},
  year={2026},
  note={To appear}
}

@inproceedings{newcombe2011kinectfusion,
  title={{KinectFusion}: Real-Time Dense Surface Mapping and Tracking},
  author={Newcombe, Richard A. and Izadi, Shahram and Hilliges, Otmar and Molyneaux, David and Kim, David and Davison, Andrew J. and Kohli, Pushmeet and Shotton, Jamie and Hodges, Steve and Fitzgibbon, Andrew},
  booktitle={Proc. IEEE Int. Symp. Mixed Augmented Reality},
  pages={127--136},
  year={2011},
  doi={10.1109/ISMAR.2011.6092378}
}

@inproceedings{keetha2024splatam,
  title={{SplaTAM}: Splat, Track \& Map {3D} Gaussians for Dense {RGB-D} {SLAM}},
  author={Keetha, Nikhil and Karhade, Jay and Jatavallabhula, Krishna Murthy and Yang, Gengshan and Scherer, Sebastian and Ramanan, Deva and Luiten, Jonathon},
  booktitle={Proc. IEEE/CVF Conf. Comput. Vis. Pattern Recognit.},
  pages={21357--21366},
  year={2024}
}

@inproceedings{huang2024zeroshape,
  title={{ZeroShape}: Regression-based Zero-shot Shape Reconstruction},
  author={Huang, Zixuan and Stojanov, Stefan and Thai, Anh and Jampani, Varun and Rehg, James M.},
  booktitle={Proc. IEEE/CVF Conf. Comput. Vis. Pattern Recognit.},
  pages={10061--10071},
  year={2024}
}

@inproceedings{downs2022gso,
  title={Google Scanned Objects: A High-Quality Dataset of {3D} Scanned Household Items},
  author={Downs, Laura and Francis, Anthony and Koenig, Nate and others},
  booktitle={Proc. IEEE Int. Conf. Robot. Autom.},
  pages={2553--2560},
  year={2022}
}

@inproceedings{wu2023omniobject3d,
  title={{OmniObject3D}: Large-Vocabulary {3D} Object Dataset for Realistic Perception, Reconstruction and Generation},
  author={Wu, Tong and Zhang, Jiarui and Fu, Xiao and others},
  booktitle={Proc. IEEE/CVF Conf. Comput. Vis. Pattern Recognit.},
  pages={803--814},
  year={2023}
}

@inproceedings{recgen2026,
  title={Reconstruction by Generation: {3D} Multi-Object Scene Reconstruction from Sparse Observations},
  author={Zadaianchuk, Andrii and Barcellona, Leonardo and Schuenemann, Lennard and others},
  booktitle={Proc. Eur. Conf. Comput. Vis.},
  year={2026}
}

@inproceedings{burde2026graspbenchmark,
  title={Benchmarking the Effects of Object Pose Estimation and Reconstruction on Robotic Grasping Success},
  author={Burde, Varun and Burget, Pavel and Sattler, Torsten},
  booktitle={Proc. IEEE Int. Conf. Robot. Autom.},
  year={2026},
}

@article{anand2026scout,
  title={Which Reconstruction Model Should a Robot Use? Routing Image-to-{3D} Models for Cost-Aware Robotic Manipulation},
  author={Anand, Akash and Agarwal, Aditya and Kaelbling, Leslie Pack},
  journal={arXiv:2603.27797},
  year={2026}
}

@article{scenecomplete,
  title={{SceneComplete}: Open-World {3D} Scene Completion in Cluttered Real World Environments for Robot Manipulation},
  author={Agarwal, Aditya and Singh, Gaurav and Sen, Bipasha and Lozano-Perez, Tomas and Kaelbling, Leslie Pack},
  journal={IEEE Robot. Autom. Lett.},
  volume={11},
  number={1},
  pages={482--489},
  year={2026},
  doi={10.1109/LRA.2025.3630884}
}

@article{carion2025sam3,
  title={{SAM 3}: Segment Anything with Concepts},
  author={Carion, Nicolas and Gustafson, Laura and Hu, Yuan-Ting and Debnath, Shoubhik and Hu, Ronghang and Suris, Didac and Ryali, Chaitanya and Alwala, Kalyan Vasudev and Khedr, Haitham and Huang, Andrew and Lei, Jie and Ma, Tengyu and Guo, Baishan and Kalla, Arpit and Marks, Markus and Greer, Joseph and Wang, Meng and Sun, Peize and R{\"a}dle, Roman and Afouras, Triantafyllos and Mavroudi, Effrosyni and Xu, Katherine and Wu, Tsung-Han and Zhou, Yu and Momeni, Liliane and Hazra, Rishi and Ding, Shuangrui and Vaze, Sagar and Porcher, Francois and Li, Feng and Li, Siyuan and Kamath, Aishwarya and Cheng, Ho Kei and Doll{\'a}r, Piotr and Ravi, Nikhila and Saenko, Kate and Zhang, Pengchuan and Feichtenhofer, Christoph},
  journal={arXiv:2511.16719},
  year={2025}
}

@inproceedings{sam3d2025,
  title={{SAM 3D}: {3Dfy} Anything in Images},
  author={Chen, Xingyu and Chu, Fu-Jen and Gleize, Pierre and others},
  booktitle={Proc. IEEE/CVF Conf. Comput. Vis. Pattern Recognit.},
  year={2026}
}

@article{centergrasp2024,
  title={{CenterGrasp}: Object-Aware Implicit Representation Learning for Simultaneous Shape Reconstruction and {6-DoF} Grasp Estimation},
  author={Chisari, Eugenio and Heppert, Nick and Welschehold, Tim and Burgard, Wolfram and Valada, Abhinav},
  journal={IEEE Robot. Autom. Lett.},
  year={2024}
}

@inproceedings{zerograsp2025,
  title={{ZeroGrasp}: Zero-Shot Shape Reconstruction Enabled Robotic Grasping},
  author={Iwase, Shun and Irshad, Muhammad Zubair and Liu, Katherine and others},
  booktitle={Proc. IEEE/CVF Conf. Comput. Vis. Pattern Recognit.},
  pages={17405--17415},
  year={2025}
}

@inproceedings{Wang_2019_CVPR_NOCS,
  title={Normalized Object Coordinate Space for Category-Level 6D Object Pose and Size Estimation},
  author={Wang, He and Sridhar, Srinath and Huang, Jingwei and Valentin, Julien and Song, Shuran and Guibas, Leonidas J.},
  booktitle={Proc. IEEE/CVF Conf. Comput. Vis. Pattern Recognit.},
  year={2019}
}

@inproceedings{todorov2012mujoco,
  title={{MuJoCo}: A Physics Engine for Model-Based Control},
  author={Todorov, Emanuel and Erez, Tom and Tassa, Yuval},
  booktitle={Proc. IEEE/RSJ Int. Conf. Intell. Robots Syst.},
  pages={5026--5033},
  year={2012},
  doi={10.1109/IROS.2012.6386109}
}

@inproceedings{nvisii,
  title={NViSII: A Scriptable Tool for Photorealistic Image Generation},
  author={Morrical, Nathan and Tremblay, Jonathan and Lin, Yunzhi and others},
  booktitle={ICLR Workshop on Synthetic Data Generation},
  year={2021}
}

@inproceedings{kaskman2019homebreweddb,
  author    = {Roman Kaskman and Sergey Zakharov and Ivan Shugurov and Slobodan Ilic},
  title     = {{HomebrewedDB}: {RGB-D} Dataset for {6D} Pose Estimation of {3D} Objects},
  booktitle = {Proc. IEEE/CVF Int. Conf. Comput. Vis. Workshops},
  year      = {2019}
}

@article{oquab2024dinov2,
  title={{DINOv2}: Learning Robust Visual Features without Supervision},
  author={Oquab, Maxime and Darcet, Timoth{\'e}e and Moutakanni, Th{\'e}o and Vo, Huy V. and Szafraniec, Marc and Khalidov, Vasil and others},
  journal={Trans. Mach. Learn. Res.},
  year={2024}
}

@article{simeoni2025dinov3,
  title={{DINOv3}},
  author={Sim{\'e}oni, Oriane and Vo, Huy V. and Seitzer, Maximilian and Baldassarre, Federico and Oquab, Maxime and Jose, Cijo and others},
  journal={arXiv:2508.10104},
  year={2025}
}

@inproceedings{hu2022lora,
  title={{LoRA}: Low-Rank Adaptation of Large Language Models},
  author={Hu, Edward J. and Shen, Yelong and Wallis, Phillip and Allen-Zhu, Zeyuan and Li, Yuanzhi and Wang, Shean and Wang, Lu and Chen, Weizhu},
  booktitle={Proc. Int. Conf. Learn. Represent.},
  year={2022}
}

@inproceedings{kashyap2025singleview,
  author={Kashyap, Abhishek and Yang, Yuxuan and Andreasson, Henrik and Stoyanov, Todor},
  title={Single-View Shape Completion for Robotic Grasping in Clutter},
  booktitle={Proc. 13th Int. Conf. Robot Intell. Technol. Appl.},
  series={Lect. Notes Netw. Syst.},
  publisher={Springer},
  address={London, United Kingdom},
  month=dec,
  year={2025}
}

@inproceedings{huang2025midi,
  title={{MIDI}: Multi-Instance Diffusion for Single Image to {3D} Scene Generation},
  author={Huang, Zehuan and Guo, Yuan-Chen and An, Xingqiao and Yang, Yunhan and Li, Yangguang and Zou, Zi-Xin and Liang, Ding and Liu, Xihui and Cao, Yan-Pei and Sheng, Lu},
  booktitle={Proc. IEEE/CVF Conf. Comput. Vis. Pattern Recognit.},
  pages={23646--23657},
  year={2025}
}

@inproceedings{meng2026scenegen,
  title={{SceneGen}: Single-Image {3D} Scene Generation in One Feedforward Pass},
  author={Meng, Yanxu and Wu, Haoning and Zhang, Ya and Xie, Weidi},
  booktitle={Proc. Int. Conf. 3D Vis.},
  year={2026}
}

@inproceedings{nolte2025ready,
  title={Is Single-View Mesh Reconstruction Ready for Robotics?},
  author={Nolte, Frederik and Geiger, Andreas and Sch{\"o}lkopf, Bernhard and Posner, Ingmar},
  booktitle={Proc. British Mach. Vis. Conf.},
  year={2026},
  note={To appear}
}

@inproceedings{zhu2022niceslam,
  title={{NICE-SLAM}: Neural Implicit Scalable Encoding for {SLAM}},
  author={Zhu, Zihan and Peng, Songyou and Larsson, Viktor and Xu, Weiwei and Bao, Hujun and Cui, Zhaopeng and Oswald, Martin R. and Pollefeys, Marc},
  booktitle={Proc. IEEE/CVF Conf. Comput. Vis. Pattern Recognit.},
  year={2022}
}

@inproceedings{sucar2020nodeslam,
  title={{NodeSLAM}: Neural Object Descriptors for Multi-View Shape Reconstruction},
  author={Sucar, Edgar and Wada, Kentaro and Davison, Andrew},
  booktitle={Proc. Int. Conf. 3D Vis.},
  year={2020}
}

@inproceedings{wang2021dspslam,
  title={{DSP-SLAM}: Object Oriented {SLAM} with Deep Shape Priors},
  author={Wang, Jingwen and R{\"u}nz, Martin and Agapito, Lourdes},
  booktitle={Proc. Int. Conf. 3D Vis.},
  year={2021}
}

@inproceedings{kong2023vmap,
  title={{vMAP}: Vectorised Object Mapping for Neural Field {SLAM}},
  author={Kong, Xin and Liu, Shikun and Taher, Marwan and Davison, Andrew J.},
  booktitle={Proc. IEEE/CVF Conf. Comput. Vis. Pattern Recognit.},
  year={2023}
}

@inproceedings{chabal2025online,
  title={Online {3D} Scene Reconstruction Using Neural Object Priors},
  author={Chabal, Thomas and Chen, Shizhe and Ponce, Jean and Schmid, Cordelia},
  booktitle={Proc. Int. Conf. 3D Vis.},
  year={2025}
}

@article{yao2025cast,
  title={{CAST}: Component-Aligned {3D} Scene Reconstruction from an {RGB} Image},
  author={Yao, Kaixin and Zhang, Longwen and Yan, Xinhao and Zeng, Yan and Zhang, Qixuan and Yang, Wei and Xu, Lan and Gu, Jiayuan and Yu, Jingyi},
  journal={ACM Trans. Graph.},
  volume={44},
  number={4},
  year={2025},
  doi={10.1145/3730841}
}

@inproceedings{guo2024physcomp,
  title={Physically Compatible {3D} Object Modeling from a Single Image},
  author={Guo, Minghao and Wang, Bohan and Ma, Pingchuan and Zhang, Tianyuan and Owens, Crystal Elaine and Gan, Chuang and Tenenbaum, Joshua B. and He, Kaiming and Matusik, Wojciech},
  booktitle={Adv. Neural Inf. Process. Syst.},
  year={2024}
}

@inproceedings{chen2024atlas3d,
  title={{Atlas3D}: Physically Constrained Self-Supporting Text-to-{3D} for Simulation and Fabrication},
  author={Chen, Yunuo and Xie, Tianyi and Zong, Zeshun and Li, Xuan and Gao, Feng and Yang, Yin and Wu, Ying Nian and Jiang, Chenfanfu},
  booktitle={Adv. Neural Inf. Process. Syst.},
  year={2024}
}

@article{xiang2026realtosim,
  title={Real-to-Sim for Highly Cluttered Environments via Physics-Consistent Inter-Object Reasoning},
  author={Xiang, Tianyi and Cao, Jiahang and Guo, Sikai and Zhao, Guoyang and Luo, Andrew F. and Ma, Jun},
  journal={IEEE Robot. Autom. Lett.},
  volume={11},
  number={7},
  pages={8544--8551},
  year={2026},
  doi={10.1109/LRA.2026.3699238}
}

@inproceedings{ni2024phyrecon,
  title={{PhyRecon}: Physically Plausible Neural Scene Reconstruction},
  author={Ni, Junfeng and Chen, Yixin and Jing, Bohan and Jiang, Nan and Wang, Bin and Dai, Bo and Li, Puhao and Zhu, Yixin and Zhu, Song-Chun and Huang, Siyuan},
  booktitle={Adv. Neural Inf. Process. Syst.},
  year={2024}
}

@article{besl1992registration,
  title={A Method for Registration of {3-D} Shapes},
  author={Besl, Paul J. and McKay, Neil D.},
  journal={IEEE Trans. Pattern Anal. Mach. Intell.},
  volume={14},
  number={2},
  pages={239--256},
  year={1992},
  doi={10.1109/34.121791}
}

@inproceedings{tatarchenko2019what,
  title={What Do Single-View {3D} Reconstruction Networks Learn?},
  author={Tatarchenko, Maxim and Richter, Stephan R. and Ranftl, Ren{\'e} and Li, Zhuwen and Koltun, Vladlen and Brox, Thomas},
  booktitle={Proc. IEEE/CVF Conf. Comput. Vis. Pattern Recognit.},
  year={2019}
}

@misc{anthropic2026fable51,
  title={Introducing {Claude Fable 5.1} and {Claude Mythos 5.1}},
  author={{Anthropic}},
  howpublished={\url{https://www.anthropic.com/claude-fable-and-mythos-5-1}},
  year={2026},
  month=sep,
  note={Accessed: 2026-09-14}
}

@IEEEtranBSTCTL{IEEEBSTcontrol,
  CTLuse_forced_etal = {yes},
  CTLmax_names_forced_etal = {6},
  CTLnames_show_etal = {1}
}

@misc{openai2026astra,
  title={{GPT-6 Astra}: A New Generation of Intelligence},
  author={{OpenAI}},
  howpublished={\url{https://openai.com/index/gpt-6-astra/}},
  year={2026},
  month=sep,
  note={Accessed: 2026-09-14}
}

\end{document}